\documentclass[letterpaper, 10 pt, conference]{ieeeconf}  %

\IEEEoverridecommandlockouts                              %

\usepackage{amsmath} %
\usepackage{mathtools}
\usepackage{amsfonts}
\usepackage{algorithm}
\usepackage{algpseudocode}
\usepackage{tablefootnote}
\usepackage{graphicx}
\usepackage{makecell}
\usepackage[bb=dsserif]{mathalpha} %
\usepackage{bm}
\usepackage{xcolor}
\usepackage{caption}
\usepackage{subcaption}
\usepackage{siunitx}
\usepackage{gensymb}
\usepackage{multirow}
\usepackage{hyperref}
\usepackage[noadjust]{cite}

\usepackage{color}
\def\eqref#1{(\ref{#1})}

\newcommand*\phantomrel[1]{\mathrel{\phantom{#1}}}

\makeatletter
\newcommand{\StatexIndent}[1][3]{%
    \setlength\@tempdima{\algorithmicindent}%
    \Statex\hskip\dimexpr#1\@tempdima\relax}
\algdef{S}[IF]{IfNoThen}[1]{\algorithmicif\ #1}%
\makeatother

\DeclarePairedDelimiterX{\infdivx}[2]{(}{)}{%
    #1\;\delimsize\|\;#2%
}

\newcommand{\pipre}{\pi_{\text{pre}}}
\newcommand{\pipost}{\pi_{\text{post}}}
\newcommand{\apre}{a_{\text{pre}}}

\title{\LARGE \bf
Pre- and post-contact policy decomposition for non-prehensile manipulation with zero-shot sim-to-real transfer
}

\author{Minchan Kim$^{1}$, Junhyek Han$^{1}$, Jaehyung Kim$^{1}$, and Beomjoon Kim$^{1}$%
\thanks{$^{1}$
        Kim Jaechul Graduate School of AI at KAIST, 
        {\tt\small\{minchan21, junhyek.han, kimjaehyung, beomjoon.kim\}@kaist.ac.kr}}%
}

\let\oldtwocolumn\twocolumn
\renewcommand\twocolumn[1][]{%
    \oldtwocolumn[{#1}{
    \begin{center}
           \includegraphics[width=\textwidth]{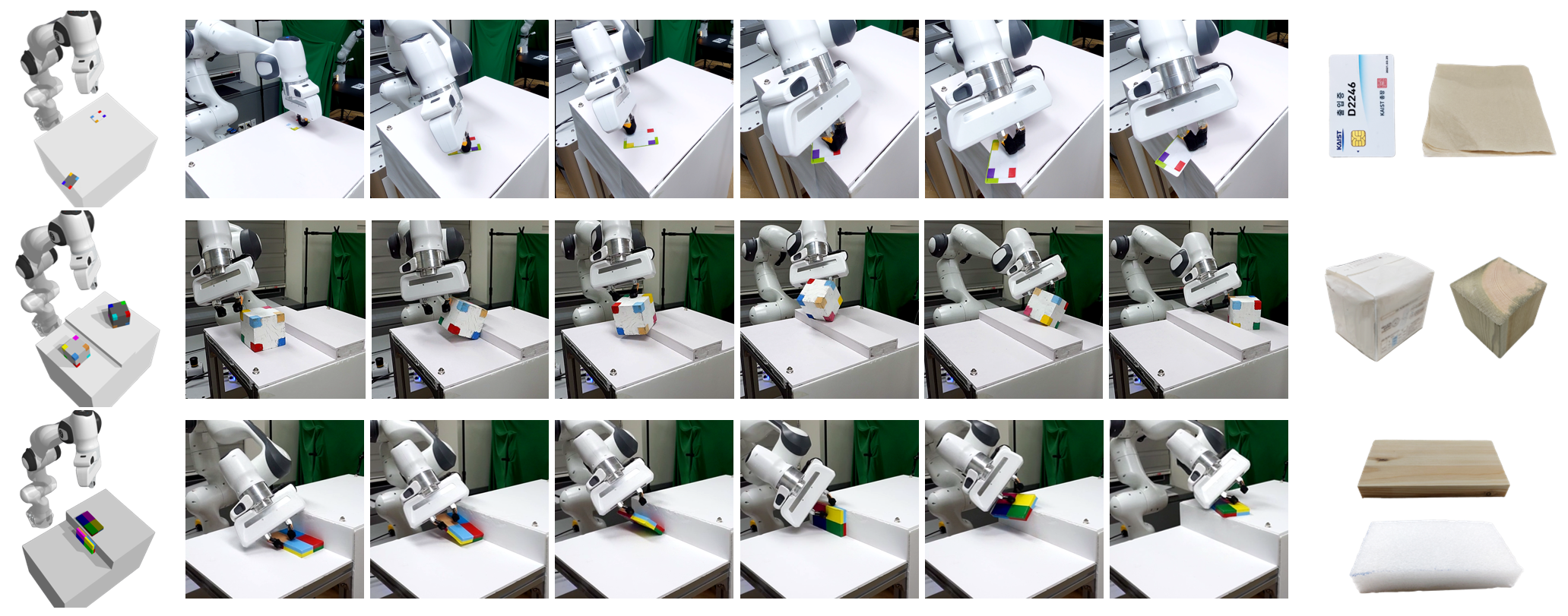}
           \captionof{figure}{Illustration of our tasks. The first column shows the initial robot and object poses and the desired object poses in the dark. The third column shows a subset of objects we manipulate in each domain. The middle column shows the manipulation motions. {\bf First row}: the card is too flat to be grasped, so the robot must use dragging and re-orientation. {\bf Second row}: the box is too large to be grasped, so the robot must push it to the bump, tumble it over, and re-orient it. {\bf Third row}: the wall to the object's right is blocking all feasible grasps. The robot must first lift it up against the wall, drag it to the top, and then finally give a little push to move it to the target location. In the last two tasks, notice how the robot must both overcome and exploit the environmental contact to manipulate the object.}
           \label{fig:overall_illustration}
        \end{center}
    }]
}
\begin{document}

\maketitle
\thispagestyle{empty}
\pagestyle{empty}

\begin{abstract}
We present a system for non-prehensile manipulation that require a significant number of contact mode transitions and the use of environmental contacts to successfully manipulate an object to a target location. Our method is based on deep reinforcement learning which, unlike state-of-the-art planning algorithms, does not require apriori 
knowledge of the physical parameters of the object or environment such as friction coefficients or centers of mass. The planning time is reduced to the simple feed-forward prediction time on a neural network. We propose a computational structure, action space design, and curriculum learning scheme that facilitates efficient exploration and sim-to-real transfer. In challenging real-world non-prehensile manipulation tasks, we show that our method can generalize over different objects, and succeed even for novel objects not seen
during training. Project website: \url{https://sites.google.com/view/nonprenehsile-decomposition}
\end{abstract}

\section{Introduction}
Humans possess remarkable skills in non-prehensile manipulation techniques such as pulling, pushing, dragging, and tumbling, which enables them to handle objects that are hard or impossible to grasp. In contrast, most robots are only capable of performing pick-and-place and fail in situations where grasping is not an option due to physical or geometric limitations. Our objective is to equip robots with the ability to manipulate objects even in such circumstances, using a basic gripper and an RGB camera as shown in Figure~\ref{fig:overall_illustration}. This is a challenging problem that involves planning a sequence of combined robot and object movements while accounting for the contact interactions among an object, robot, and environment. Since the robot is equipped with a simple gripper, it must leverage the gravity and environment's geometry to perform the task effectively. 

The state-of-the-art algorithms for such problems are planning algorithms that use tree search or trajectory optimization based on an analytical physics model~\cite{gilwoo2015iros,jenny2013icra,cheng2022contact, posa2014direct, mordatch2012contact,bernardo2020rss}. However, their applicability in the real world is limited by several factors. Firstly, their search space is extremely complex, involving both discrete contact activations and continuous robot and object motions, with different sub-manifolds defined by different motion constraints. This makes planners too slow to use in practice. Secondly, they require accurate analytical contact models of the world, which is non-trivial to define as rigid-body dynamics are often insufficient~\cite{lynchcontact}. Moreover, even if we \emph{can} define the governing equation of contacts, extracting the physical parameters for instantiating the equation, such as the center of mass or friction coefficient, from high-dimensional RGB images is a challenging problem.

Because of these limitations, they have only been applied in a controlled setup where all necessary information about the object, such as inertial parameters or pose, is known. To achieve reasonable planning time, they often rely on strong assumptions about contact interactions, robot motions, and transition dynamics, such as quasi-static dynamics, predefined contact modes (slipping, static, rolling, etc.), and predefined contact locations in the environment, object, and robot~\cite{cheng2022contact, miyazawaISATP2005, gilwoo2015iros,mordatch2012contact}.

We propose an alternative solution for non-prehensile manipulation that utilizes deep reinforcement learning (RL). By directly learning the representation that implicitly encodes the necessary information from sensory data, our approach eliminates the need to extract difficult-to-observe object inertial parameters. At the expense of offline training, the online computation is reduced to feed-forward predictions from a neural network rather than searching in a complex space. Notably, our approach does not make assumptions about contact interactions and robot or object motions, making it more versatile than traditional planning algorithms.

Like the recent approaches for contact-rich tasks such as in-hand manipulation~\cite{chen2022system,chen2022visual,andrychowicz2020learning} and locomotion~\cite{hwangbo2019science,lee2020learning}, we use a simulator to train a policy and then transfer it to the real world. In this scheme, there are two primary challenges that we need to address for non-prehensile manipulation: exploration and the sim-to-real gap. Exploration in non-prehensile manipulation is challenging because unlike locomotion or in-hand manipulation where the object or the ground is in close proximity to make contact, the object is not in a position where we can directly make contact, especially in the initial state. So, if we randomly explore, there is zero chance of sampling an action that contacts the object~\cite{TAMPsurvey}. The sim-to-real gap arises from the inaccuracies in the physics engine and modeling. We found that the hardware of the widely-used collaborative robots such as Franka Emika Panda is based on industrial robots and has a large gear ratio with intricate joint friction that is difficult to model accurately.

\begin{figure}[t]
    \centering

    \begin{subfigure}[b]{0.26\columnwidth}
         \includegraphics[width=\textwidth]{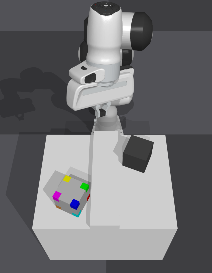}
         \caption{EE-ABOVE}
         \label{fig:EE-ABOVE}
     \end{subfigure}
    \hfil
     \begin{subfigure}[b]{0.42\columnwidth}
         \includegraphics[width=\textwidth]{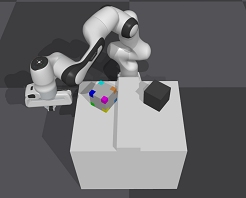}
         \caption{EE-AT-RIGHT}
         \label{fig:EE-AT-RIGHT}
     \end{subfigure}
    
    \caption{(a) EE-ABOVE: The end-effector is placed above the object. (b) EE-AT-RIGHT: The end-effector is placed on the right side of the object. The black cube denotes the goal pose. The robot should contact the object on the right to manipulate it to the goal, and initially contacting it on top of the object makes exploration extremely challenging as it would have to break the contract entirely and re-make it on the right.}
    \label{fig:claim_1_fig}
\end{figure}

Our primary contribution is the design of a computational structure, action space, and learning curriculum that solves both of these challenges. For the exploration problem, one naive approach would be to define a contact-inducing reward that rewards the robot if its hand gets closer to the object. However, for non-prehensile manipulation, not all contacts are equal: some initial contacts are more crucial as it leads to more promising state space in the post-contact phase, as shown in Figure 2. Unfortunately, such a reward would make the robot contact the closest point on the object, and we empirically found this approach too sensitive to the initial system state.

Instead, we make a key observation that in general, any non-prehensile manipulation can be divided into two stages: the \emph{pre-contact} and \emph{post-contact} stages. In the pre-contact stage, the robot primarily focuses on finding a promising initial contact on the object that would lead to successful manipulation. In the post-contact phase, the robot finds a sequence of forces to apply to the object that successfully moves it to the target location. In this latter stage, the robot will mostly be in contact with the object — however, it might purposefully break a contact to exploit gravity and environmental contact or simply change its contact point to exert a force from a different direction. 

Based on this observation, we propose a computational structure
that facilitates effective exploration, where we use two different policies for each stage: pre-contact and post-contact policies, each having different action spaces. The \emph{pre-contact policy}’s action is defined by the point on the object and the gripper that together defines the contact position, along with the 
gripper's orientation and gripper width at the contact. This guarantees contact with every action and enables the robot to explore the space of contacts more efficiently than say, using a joint torque or SE(3) end-effector pose. We train this policy first in a simulator where we have accurate object shape information and then transfer it to the real world by using a student-teacher training scheme~\cite{chen2020learning}.

Because the pre-contact policy involves just position control, there is a negligible sim-to-real gap. The design of the post-contact policy’s action space, on the other hand, requires more consideration as it involves contact interaction with the object. In locomotion and in-hand manipulation, the typical approach is to learn a policy to generate the next target \emph{joint position}, and use an analytical joint position controller~\cite{hwangbo2019science,chen2022system,andrychowicz2020learning}. On the other hand, for our tasks with natural motion constraints\footnote{In the context of hybrid position-force control} and continual contacts, learning a policy that generates the target \emph{end-effector pose} along with control gain parameters for an operational space controller (OSC) has shown to have the best sample efficiency among different action spaces~\cite{vices2019}.

Unfortunately, OSC requires an accurate robot model~\cite{nakanishi2008operational}, but there are several factors hindering accurate modeling; using a system identification helps, but we found it to be insufficient. For instance, the wrist joint on Panda has large friction in one direction but not so much in the opposite direction, and such intricacy cannot be captured with the parameters available in a typical physics simulator. Furthermore, since OSC uses a Cartesian space error to compute joint torques unless the end-effector moves exactly the same way for the given torque in simulation and the real world, there would be a large joint position gap.

We instead propose a different action space. Just like in~\cite{vices2019}, we predict the end-effector target pose. However, instead of predicting the gains for an OSC, we predict the gains for each joint and use inverse kinematics to solve for the joint position target. Then, we use a joint position controller to realize that joint position using the predicted controller gains. We found that using the joint position controller transfers much more effectively to the real world than OSC because the controller errors are now in the joint space rather than Cartesian space which you can correct by moving the respective joint.

Another factor we need to consider when transferring a policy to the real world is the violation of joint velocity and torque limits. The real robot has safety measures that prevent the robot to move above a certain speed and contact force. To account for this, we can terminate the episode if the robot violates them during learning, as in~\cite{chen2022visual}. However, we found this makes learning difficult as it explores too conservatively, and fails to learn a policy in some of our domains. So, we introduce an action scaling curriculum that allows the robot to explore aggressively at the beginning of the learning but slowly decrease the maximum magnitude of the action so that the velocity and torque limits are met. This facilitates both efficient exploration and sim-to-real transfer.

As in previous works~\cite{chen2022system,cheng2022contact,hwangbo2019science,andrychowicz2020learning}, we make significant use of domain randomization to close the sim-to-real gap. 
We found that, with an end-to-end system, it is difficult to determine where the sim-to-real gap lies since the system is not modularized. So we design a modular system where the perception module takes a single RGB and outputs the 2D object key points. The policy takes the 2D key points and the proprioceptive sensor data to produce an action. We train our key point detector, pre-contact, and post-contact policies entirely in simulation and transfer them to the real world.

We apply our method to three challenging non-prehensile manipulation problems where environment or object geometry renders grasping impossible, as shown in Figure~\ref{fig:overall_illustration}. The robot needs to make use of both intrinsic and extrinsic contacts to manipulate objects that are not directly graspable. We show that not only can our method handle objects seen in training, but also new objects with different friction coefficients, centers of mass, and densities that are outside of the domain randomization range.

\section{Related work}
\subsection{Planning through contacts}
There are several methods for planning through contacts that can be applied 
to our tasks. The first class of algorithms is
optimization-based methods \cite{mordatch2012contact,posa2014direct,moura2022non} 
which analytically models the robot kinematic and dynamics constraints, such as joint limits, and contact constraints, such as friction cone constraints, as constraints in an optimization problem, and uses gradient-based techniques to find a trajectory. To handle the hybrid search space and discontinuous dynamics, some methods smoothen the contact mode decisions~\cite{mordatch2012contact} or use complementarity constraints~\cite{posa2014direct,moura2022non}. While these algorithms can compute impressive motions with many contact mode changes, they also tend to output motions that are unrealistic because it is difficult to satisfy the constraints and to accurately model contacts. To our knowledge, these methods have scarcely been demonstrated on a real robot, if not never.

Another class of planning algorithms is based on graph search~\cite{zito2012two,miyazawaISATP2005,gilwoo2015iros,cheng2022contact,liang2022learning}. 
To handle discontinuous dynamics and hybrid search space, these methods typically construct a graph
where each node encodes the contact mode and states of the object and robot, and each edge encodes the motion between two states if the transition is feasible. To compute the motion on each edge, either a
pre-defined motion primitive~\cite{zito2012two} or an additional planner is used that accounts for dynamics constraints imposed by the contact mode and state~\cite{miyazawaISATP2005,cheng2022contact,liang2022learning,gilwoo2015iros}. In general, these methods output more physically realistic motions than optimization-based algorithms and have been demonstrated on real robots~\cite{cheng2022contact,moura2022non,liang2022learning}. However, they make strong assumptions such as quasi-static assumption or a pre-defined set of primitives or contact modes, and are limited to tasks that can be solved using simple motion with very few or no contact mode changes. 

Because all these algorithms must compute a plan in a hybrid search space with discontinuous dynamics, they are too slow to use in practice. Furthermore, except~\cite{liang2022learning}, which predicts whether a motion
primitive and its parameters will succeed from a segmented point cloud of the scene, all methods depend on the assumption that physical parameters such as mass and friction coefficient are known. In contrast, our approach suffers from neither of these problems. We can compute the next action by making a feed-forward prediction using a neural network, which usually takes on the order of milliseconds, and operate directly using a high dimensional camera and proprioceptive sensor rather than relying on accurate physical parameter knowledge.

\subsection{Reinforcement learning for contact-rich tasks}
RL algorithms have been successfully demonstrated for in-hand manipulation
\cite{akkaya2019solving,andrychowicz2020learning,allshire2021transferring,chen2022system,chen2022visual,dextreme2022} tasks. One key difference between these and our work is that in in-hand manipulation, the object typically starts in close proximity to the robot, whereas in our domain, we must solve the additional problem of reaching and making contact with the object.
With an exception to~\cite{chen2022system}, which has the object right beneath the hand, all systems begin with the object in hand. Like in~\cite{chen2022system}, we encourage contact between the object and the finger using a reward. However, we found this to be insufficient
as the initial contact is more crucial in our domain, and use a pre-contact policy to facilitate efficient exploration. Following \cite{allshire2021transferring}, we use object key points to represent their pose.

There are few works on RL-based non-prehensile manipulation~\cite{yuan2018rearrangement,yuan2019end,lowrey2018reinforcement,peng2018sim}.
Unlike planning-based algorithms, these methods can handle high-dimensional input data such as images, do not explicitly depend on physical parameters, and have been demonstrated in the real world. However, these methods are limited to planar pushing,  whereas planning algorithms could synthesize complex motions which incorporate multiple contact mode changes. 
Our method on the other hand not only generates complex movements, but also operates on high-dimensional sensory data without physical parameters.

The work by Zhou and Held~\cite{zhou2022learning} is close to ours in that it trains an agent to grasp an initially ungraspable object, and shows generalization over a variety of objects. 
One key difference from our work is that we focus on moving the object to the goal pose, whereas their objective is to grasp the object.  Moreover, since they reward the agent to approach the target grasp, the motion typically involves only a few mode changes. Finally, they use an OSC and must operate the policy at a very low frequency to close the sim-to-real gap. However, our method controls joint position directly and controls the robot at a much higher frequency, enabling more dexterous motions.

\section{Method}

\begin{table}[t]
    \caption{The components of state space $\mathcal{S}$ of $\pipost$.}
    \label{tab:pi-post-state}
    \centering
    \begin{tabular}{|c|c|}
        \hline
        Component & Description \\
        \hline
        $q[t] \in \mathbb{R}^{9}$ & joint and finger positions at time step $t$ \\
        $\dot{q}[t] \in \mathbb{R}^{9}$ & joint and finger velocities at time step $t$\\ 
        $u_{o}[t] \in \mathbb{R}^{2 \times 8}$ & 2D object key points at time step $t$ \\
        $u_{g} \in \mathbb{R}^{2 \times 8}$ & 2D goal object key points \\
        $T_{E}[t] \in SE(3)$ & end-effector pose at time step $t$ \\
        $a_{post}[t-1]$ & post-contact policy's action at time step $t-1$\\
        \hline
    \end{tabular}
\end{table}

\begin{figure*}[!th]
    \centering
    \includegraphics[width=0.8\textwidth]{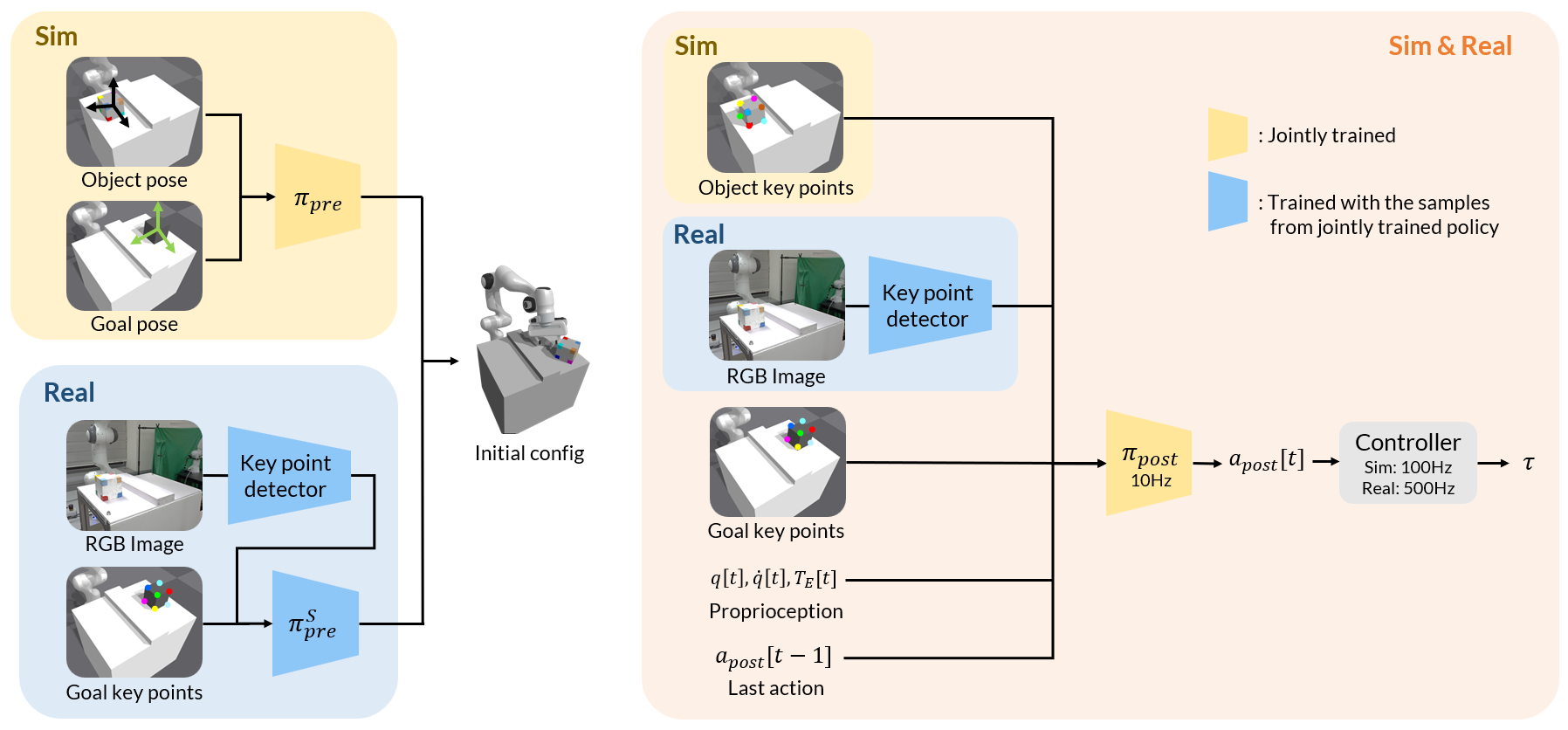}
    \caption{Modules in our system. (Top left) In the simulation, we use the ground-truth object and goal pose as inputs to $\pipre$, which defines the initial robot configuration for the post-contact policy. (Bottom left) In the real world, we use the output of the key point detector and $\pipre^S$, the student pre-contact policy. (Right) $\pipost$ in simulation uses the ground-truth 2D key points along with other inputs, while in the real world, it uses the output of the key point detector.
    }
    \label{fig:architecture}
\end{figure*}

 We are given a single RGB camera, a manipulator with a simple gripper, and a proprioceptive sensor. The target location of the object is defined using a relative transform with respect to its initial pose. We use Franka Emika Panda in this paper, but our method can be applied to other manipulators. Figure~\ref{fig:architecture} shows how different modules in our system get trained and used in simulation and the real world.

\subsection{Training $\pipre$ and $\pipost$ in a simulator}
\begin{figure}[htbp]
    \centering
    \includegraphics[width=0.4\textwidth]{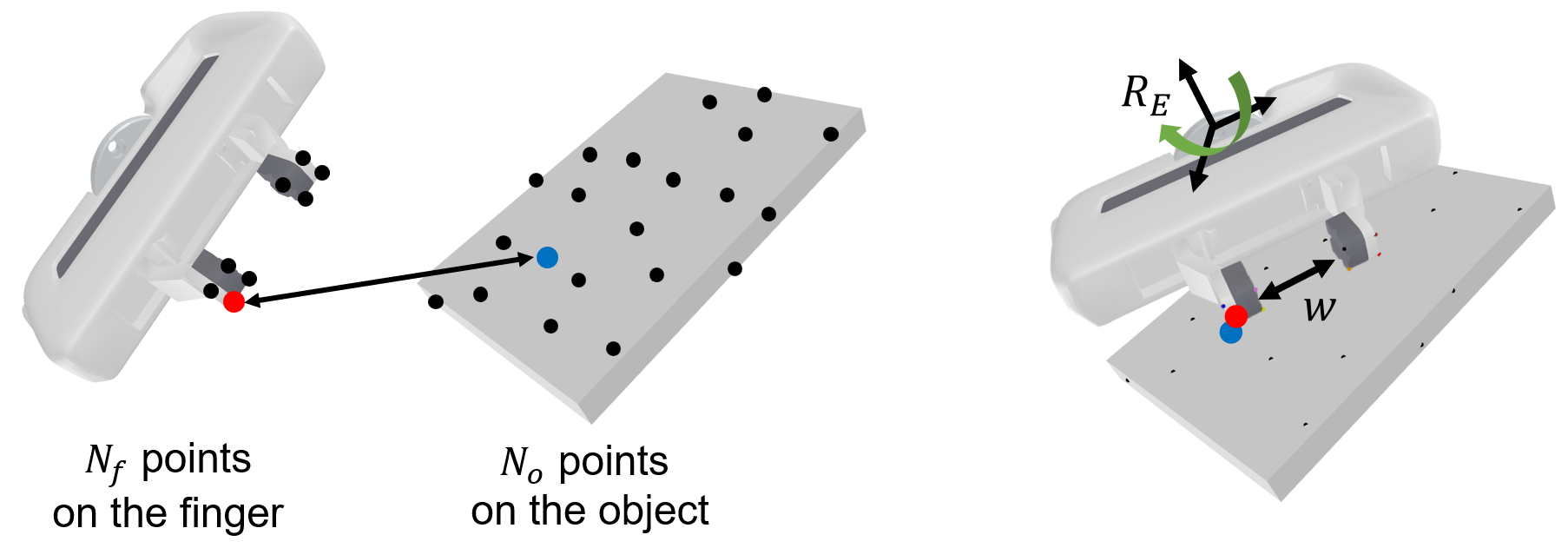}
    \caption{Illustration of how $\pipre$ makes the initial contact.}
    \label{fig:pre_contact}
\end{figure}

The state space of $\pipre$ consists of $(T_o, T_g)\in SE(3)\times SE(3)$ where $T_o$ and $T_g$ denote initial and goal object poses respectively. Its action space consists of four components: the orientation of the gripper, $R_E \in SO(3)$, its width, $l\in[0,0.04]$, a point on the gripper, $c_{f}$, and a point on the object, $c_{o}$. The space of $c_f$ is defined as the pre-defined
points on the robot gripper, and the space of $c_o$ is defined 
as points uniformly sampled from the surface of the object.
The desired end-effector position, denoted $p_c$, is defined by matching $c_f$ and $c_o$. 
See Figure~\ref{fig:pre_contact} for an illustration. We compute the joint position at $(p_c, R_E)$ using inverse kinematics (IK) solver.

We define the 3D key points as the vertices of the 3D bounding box for the target object. The state space of the post-contact policy $\pi_{post}$ consists of variables listed in Table~\ref{tab:pi-post-state}. Its action space consists of the end-effector \textit{residual} 
denoted with $\Delta T_{E}[t] \in SE(3)$, defined as the difference between the target and
the current end-effector pose, the controller gain $k_{p}[t]$ and the damping ratio $\rho[t]$ for the joint controller. The controller gain $k_{d}[t]$ is calculated by $k_{d}[t] = \rho[t] \cdot \sqrt{k_{p}[t]}$. To execute an action, we use IK to solve for the joint position target in the next step, and then run the joint position controller with the controller gains.

We train both $\pipre$ and $\pi_{post}$ 
using Proximal Policy Optimization (PPO)~\cite{schulman2017proximal} using the reward function defined as
\begin{flalign}
    r(s[t],a_{post}[t])&=\sum_{i=1}^{N_k}\frac{C_1}{\|x_{o}^{i}[t]-x_{g}^{i}\|_{2}+C_1}-\|k_{p}[t]\|_{2}&\nonumber\\
    &\phantomrel{=}+C_2\mathbb{1}\left(d[t]<\bar{d}\;\text{and}\;\theta[t]<\bar{\theta}\right)&\nonumber\\
    &\phantomrel{=}+\frac{C_3}{\|(p_{\mathrm{lf}}[t]+p_{\mathrm{rf}}[t])/2-p_{\mathrm{obj}}[t]\|_{2}}\ ,
    \label{eqn:pi-post-reward}
\end{flalign}
where $C_1, C_2$, and $C_3$ are positive constants, 
$x_{o}^{i}[t] \in \mathbb{R}^{3}$ and $x_{g}^{i} \in \mathbb{R}^{3}$ are 3D current and goal object key points, 
and $N_k$ is the number of key points on the object. The first term increases as the object get closer to the goal, where the distance is computed based on the 3D key points. 
The second term regularizes the magnitude of $k_{p}$ to induce a compliant motion. The third term is an extra reward for completing the task, in addition to matching the key points for completing the job. Here, $d[t]$ is the distance between the goal and current object position, and $\theta[t]$ is the distance between the goal and current object orientation, measured using a quaternion difference $\alpha[t]\alpha_{g}^{-1}$  between the current and goal object orientation, denoted $\alpha[t]$ and $\alpha_{g}$ respectively. Typically we set $C_2 \gg C_1$. The last term induces the end-effector to be close to the object. Here, $p_{\mathrm{obj}}$ is the position of the center of mass of the collision mesh of the object, and $p_{\mathrm{lf}}$ and $p_{\mathrm{rf}}$ are at the left and right finger joints respectively.

We train $\pipre$ and $\pipost$ jointly. Given an environment, we first create a problem by sampling a $(T_o[0], T_g)$ pair. We then use $\pipre$ to obtain its action, $\apre$. If $\apre$ yields an end-effector pose that is in a collision or is kinematically infeasible, we terminate the episode, receive $C_4<0$ as a reward, and update $\pipre$ with this reward. Otherwise, we execute $\apre$ by setting the robot at joint configuration $q_E$
given by the IK solver for $\apre$. Then, we execute $\pipost$, updating it at every $H$
time steps until the episode terminates.  In this phase, an episode can terminate because (1) the maximum episode length, $L$, has been reached, (2) the robot drops the object, or (3) the robot succeeds in the task. Once the episode terminates, we take the sum of the rewards that we have obtained during the episode and update $\pipre$. The process repeats for a desired number of iterations.

\subsection{Sim-to-real transfer}

\subsubsection{Transferring $\pipre$}
Since the state and action of $\pipre$ depend on information unavailable in the real world, we cannot use $\pipre$ directly. So, we train a student policy, $\pipre^{S}$, using $\pipre$ as the teacher policy~\cite{chen2020learning}. The student policy's action consists of $q_E\in SE(3)$, the end-effector pose, and $l$. Its state consists of the 2D current and goal object key points, $u_o[t] \in \mathbb{R}^{2 \times 8}$, and $u_g \in \mathbb{R}^{2 \times 8}$. We generate the imitation learning data for $\pipre^{S}$ by running a trained $\pipre$ in the simulator. We convert actions of $\pipre$ to $q_E$ and $l$ pair and use them as labels for $\pipre^{S}$. We also convert $T_{o}$ and $T_{g}$ into $u_o[t]$ and $u_g$ by projecting the vertices of the bounding box of the object to the image plane. We train $\pipre^{S}$ by minimizing the mean squared error (MSE). To ensure the MSE loss makes sense on the orientation, we represent the orientation of $q_{E}$ in the form of 6D representation used in \cite{zhou2019continuity}, which is just the first two columns of a rotation matrix. %

\subsubsection{Training and transferring a perception module}
We train a key point detector using synthetic images in simulation and then use
domain randomization to transfer it to the real world. Our detector takes a single RGB image as an input and outputs 2D object key points. We generate training data for the key point detector by running trained $\pipre$ and $\pipost$ in the simulator. At each time step, we take an RGB image, a segmentation mask indicating whether each pixel belongs to the robot, the object, the table, or the background, and ground-truth 2D key points projected onto an image plane. 

We adopt the key point detector network architecture from~\cite{s3k2021}, which generates $N_k$ heatmaps, one for each 2D key point, with the same size as the input image. These heatmaps represent the probability of each 2D key point's existence at each pixel. To train the network, we use ground-truth Gaussian heatmaps centered at the projected location on the image plane for each key point, as in~\cite{s3k2021}. We train the key point detector by minimizing the KL divergence between the predicted heatmaps and the ground-truth heatmaps.
To make the key point detector robust to changes in the background and the environment  when we transfer to the real world, we use momentum contrast learning~\cite{moco2020} which enables the key point detector to focus on the object rather than its surroundings. We add InfoNCE term~\cite{oord2018representation} to the KL divergence loss. To augment the images, we replaced the RGB values of the pixels belonging to the background and the table by referring to the segmentation mask. With probability 0.5, we replaced the pixel RGB values with random RGB colors, and with probability 0.5, we replaced them with textures used in \cite{jiang2022vima}.

\subsubsection{Correcting the errors in a robot model}
To close the sim-to-real gap in the robot model, we perform system identification on
the joint dynamics parameters available in the simulator, denoted $\alpha$, which includes joint friction, damping, and armature. To do this, we first collect $N_{traj}$ trajectories with the length of $\mathcal{T}$ from the real robot by setting one of the seven joints to follow a sinusoidal joint position target of the form $q^{real}[t]=A\sin(\beta  \gamma  t)$, while the remaining joints maintain their position. The amplitude $A$ and frequency $\gamma$ of the trajectory are selected randomly from the hand-tuned range where its maximum is proportional to the position and velocity limits of each joint. We then optimize $$ \min_{\alpha}\sum_{i=0}^{N_{traj}}(\sum_{t\in\mathcal{T}}(q_i^{real}[t]-q_i^{sim}[t;\alpha])^2)^{\frac{1}{2}}$$ for each joint using CMA-ES~\cite{cmaes2003}, where $q_i^{sim}[t;\alpha]$ is the joint position that would result in the simulator when we use
a joint position controller to follow the same sinusoidal trajectory with dynamics parameters set to $\alpha$. Since only one joint moves in each trajectory, we can treat all joints separately, which simplifies the optimization problem.

\subsubsection{Meeting the joint limits}
The real robot has joint velocity and torque limits. To satisfy them, we must limit the
magnitude of the end-effector residual at each time step, but
this hinders efficient exploration. We use a curriculum for the maximum end-effector residual magnitude to handle this problem. Denote the target end-effector residual magnitude that satisfies the joint limits as $\zeta^*$, which is obtained from the real robot. We begin with a large initial limit $\zeta_{o}>\zeta^*$, which gets reduced whenever the policy success rate reaches 80\%. The scale is reduced in a geometric sequence with ratio $(\zeta^*/\zeta_{o})^{\frac{1}{N_{s}}}$ where $N_{s}$ is a hyperparameter that determines the number of steps to reach the target residual magnitude from the initial magnitude.

\subsubsection{Domain randomization}
To make the policy more robust to domain difference and the noise when it is transferred to the real robot, we further train our policy with domain randomization (DR) after we train our policy with an action scaling curriculum. Below is the list of properties we applied and the detailed ranges are noted in Table~\ref{tab:DR}.
\begin{itemize}
    \item Physics properties: The friction of the table and robot and the mass of the object are randomized. We also add random noise to the commanded torque from the controller, to reflect the effects of real-world noise.
    \item Robot model: We adaptively reduce the gap between minimum and maximum values for the joint position range whenever the policy reached a success rate over a threshold.
    \item Perception: We add noise to the input of the policy to mimic the sensor noise and error of the key point detector.
\end{itemize}

\begin{table}[t]
    \caption{Domain randomization noise. $\mathcal{U}[min, max]$ denotes uniform distribution, and  $\mathcal{N}[\mu,\sigma]$ denotes Normal distribution.}
    \label{tab:DR}
    \centering
    \begin{tabular}{|c|c|}
        \hline
        Parameter & Range \\
        \hline
        Table friction & $\times\mathcal{U}[0.7,1.3]$ \\
        Robot end-effector surface friction & 
        $\times\mathcal{U}[0.9,1.1]$ \\
        Object mass & $\times\mathcal{U}[0.7,1.3]$\\
        Torque noise & $+\mathcal{N}[0.0,0.03]$ \\
        2D key point noise & $+\mathcal{N}[0.0,0.03]$ \\
        Sensor noise & $+\mathcal{N}[0.0,0.01]$ \\
        \hline
    \end{tabular}
\end{table}

\section{Experiments}
\subsection{Domain description and experiment setup}

\begin{figure*}[htb]
    \centering
    \begin{subfigure}[b]{0.24\textwidth}
         \includegraphics[width=\textwidth]{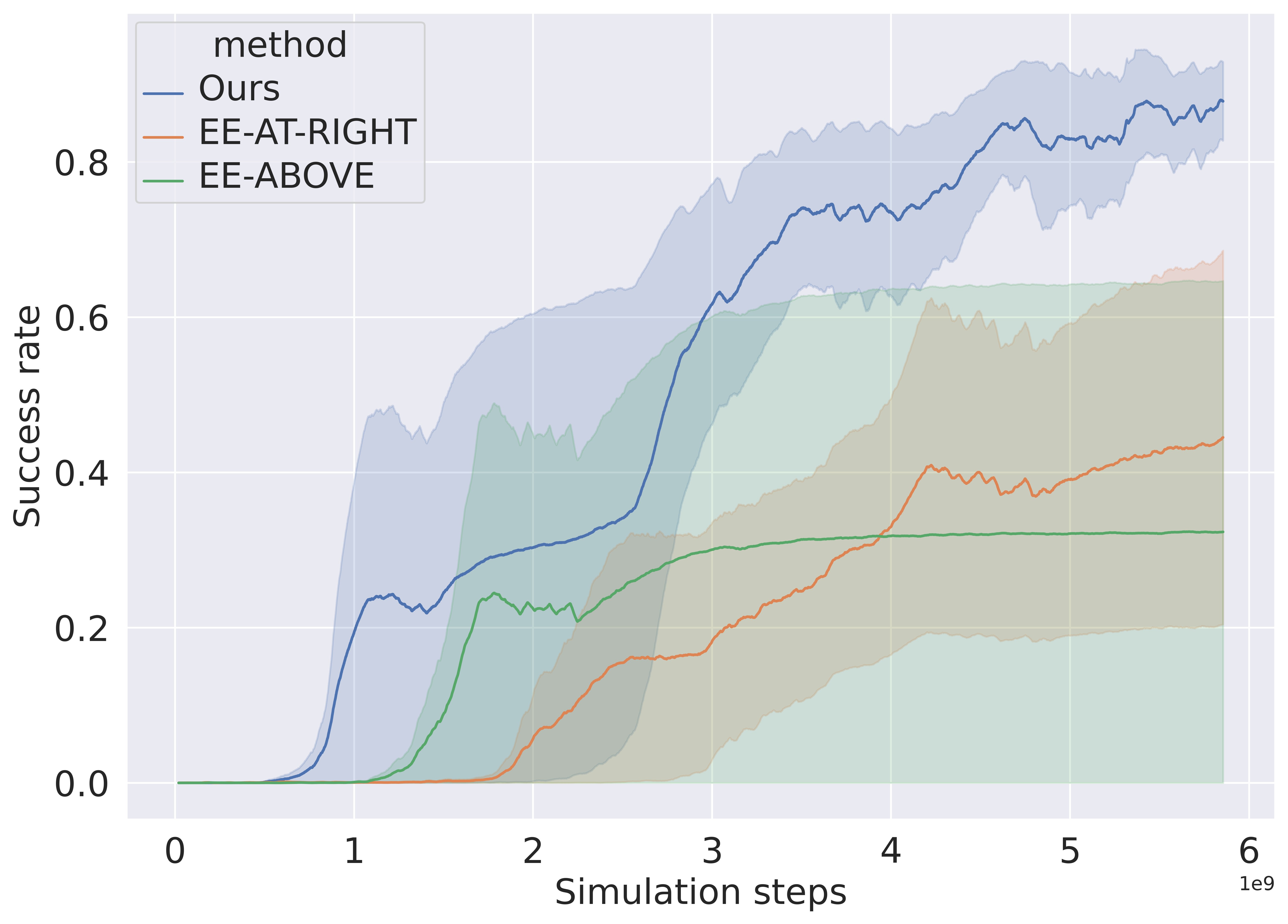} %
         \caption{Bump domain}
         \label{fig:plot1}
     \end{subfigure}
     \begin{subfigure}[b]{0.24\textwidth}
         \includegraphics[width=\textwidth]{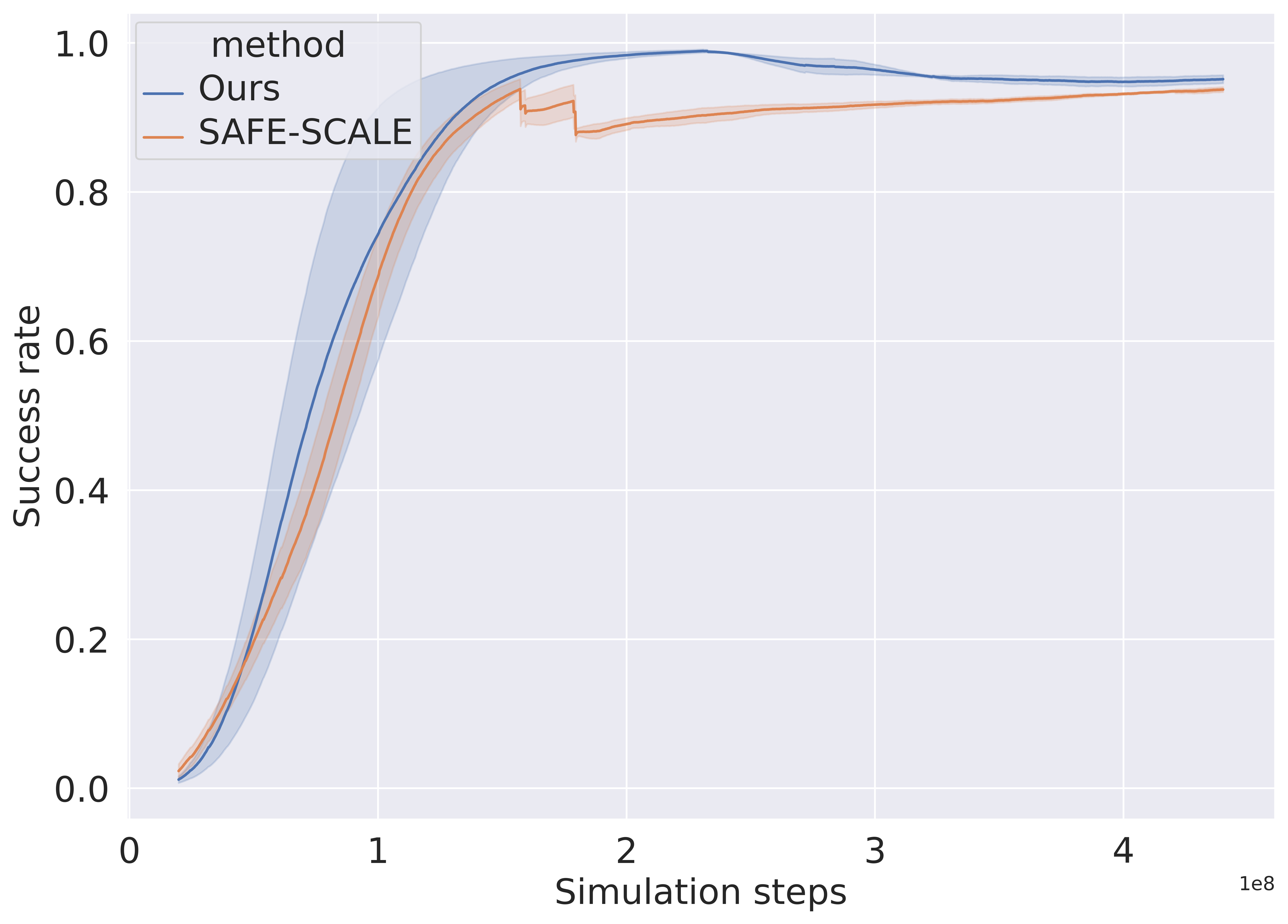}
         \caption{Card domain}
         \label{fig:plot2}
     \end{subfigure}
     \begin{subfigure}[b]{0.24\textwidth}
         \includegraphics[width=\textwidth]{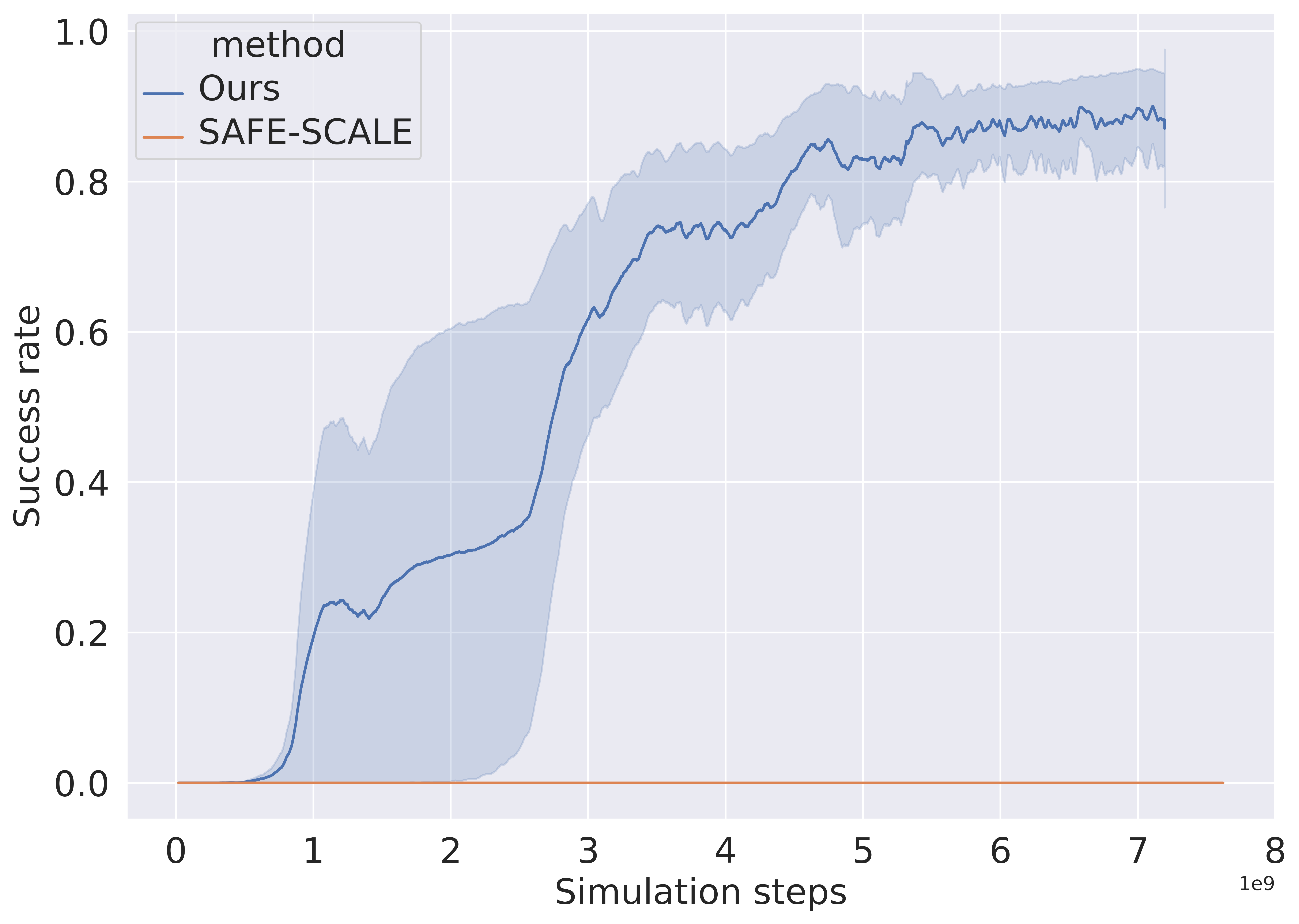}
         \caption{Bump domain}
         \label{fig:plot3}
     \end{subfigure}
     \begin{subfigure}[b]{0.24\textwidth}
         \includegraphics[width=\textwidth]{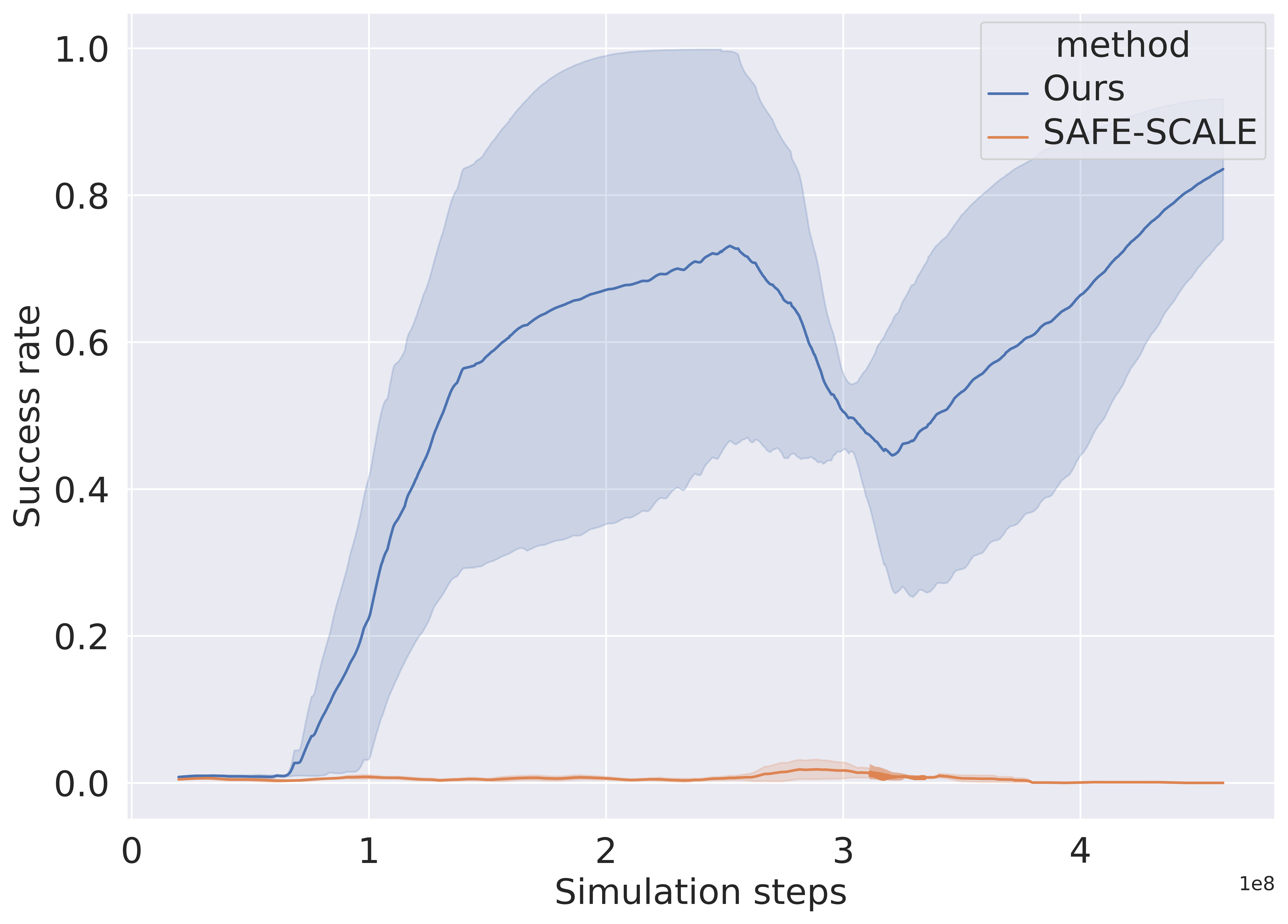}
         \caption{Wall domain}
         \label{fig:plot4}
     \end{subfigure}
     \caption{ Learning curves for (a) comparing our method with the one that does not use pre-contact policy in the bump domain, (b, c, d) comparing our method to the one without scheduling in card, bump, and wall domains, respectively. The sudden dips or temporary plateaus in the graphs are due to domain randomization and residual scaling curriculum.}
    \label{fig:result plot}
\end{figure*}

\begin{table*}[htb]
\centering
\resizebox{0.72\textwidth}{!}{%
\begin{tabular}{|c|c|c|c|c|c|c|c|c|}
\hline
Domain                & Controller & Object                    & Mass (g) & Surface & scenario 1 & scenario 2 & scenario 3 & Success rate \\ \hline
\multirow{9}{*}{Card} & Ours       &      \multirow{3}{*}{3D printed (default)} & \multirow{3}{*}{8} & \multirow{3}{*}{plastic} & 5/5 & 3/5 & 5/5  & 0.87         \\
                      & OSC        &                                             &                   &                           & 1/5 & 1/5 & 0/5 & 0.13         \\
                      & $\zeta^{*}$ &                                          &                   &                           & 5/5 & 2/5 & 4/5 & 0.73         \\ \cline{2-9}
                      & \multirow{6}{*}{Ours}       & 3D printed (wrapped)       & 8        & vinyl   & 4/5        & 1/5        & 2/5        & 0.47         \\
                      &        & Acrylic                    & \textbf{16*}      & acrylic & 3/5        & 3/5        & 5/5        & 0.73         \\ 
                      &        & Wood                       & 8        & wood    & 5/5        & 4/5        & 5/5        & 0.93         \\
                      &        & Credit card                & 5        & plastic & 5/5        & 5/5        & 4/5        & 0.93         \\
                      &        & \textbf{Tissue$\dag$}               & \textbf{1*}       & tissue  & 4/5        & 5/5        & 5/5        & 0.93         \\
                      &        & \textbf{Chocolate chip$\dag$}           & 9        & vinyl   & 3/5        & 4/5        & 5/5        & 0.87         \\ \hline
\multirow{8}{*}{Bump} & Ours       & \multirow{2}{*}{3D printed (default)}       & \multirow{2}{*}{146}      &  \multirow{2}{*}{paper}   & 5/5        & 3/5        & 5/5        & 0.87         \\
                      & OSC        &                                               &                          &     & 0/5        & 0/5        & 1/5        & 0.07         \\ \cline{2-9}
                      & \multirow{6}{*}{Ours}       & 3D printed (wrapped)       & 146      & vinyl   & 4/5        & 2/5        & 5/5        & 0.73         \\
                      &        & 3D printed (high density)  & \textbf{223*}     & paper   & 5/5        & 5/5        & 5/5        & 1.00         \\
                      &        & Wood                       & \textbf{381*}     & paper   & 5/5        & 3/5        & 4/5        & 0.80         \\
                      &        & Painted wood               & \textbf{381*}     & wood    & 2/5        & 5/5        & 3/5        & 0.67         \\
                      &        & \textbf{Sponge$\dag$}               & \textbf{30*}      & paper   & 2/5        & 5/5        & 3/5        & 0.67         \\
                      &        & \textbf{Diaper box$\dag$}                & 102     & paper   & 2/5        & 4/5        & 2/5        & 0.53         \\ \hline
\multirow{8}{*}{Wall} & Ours       & \multirow{2}{*}{3D printed (default)}       & \multirow{2}{*}{38}       & \multirow{2}{*}{paper}   & 5/5        & 3/5        & 5/5        & 0.87         \\
                      & OSC        &        &        &    &  0/5          & 0/5            &   0/5         & 0.00             \\ \cline{2-9}
                      & \multirow{6}{*}{Ours}       & 3D printed (wrapped)       & 38       & vinyl   & 5/5        & 3/5        & 5/5        & 0.87         \\
                      &        & 3D printed (high density)  & \textbf{62*}      & paper   & 5/5        & 4/5        & 4/5        & 0.87         \\
                      &        & Wood                       & \textbf{97*}      & paper   & 5/5        & 0/5        & 4/5        & 0.60         \\
                      &        & Painted wood               & \textbf{97*}      & wood    & 5/5        & 1/5        & 3/5        & 0.60         \\
                      &        & \textbf{Sponge$\dag$}               & \textbf{5*}       & paper   & 5/5        & 5/5        & 5/5        & 1.00         \\
                      &        & \textbf{Water tissue box$\dag$}         & \textbf{138*}     & paper   & 0/5        & 1/5        & 2/5        & 0.20         \\ \hline
\end{tabular}%
}
\caption{The real-world results in diverse objects for three domains. $\dag$ means the object is non-rigid, and * means the mass of the object is out of DR range. $\zeta^{*}$ means the policy is trained with the same controller and architecture as our system except that it uses fixed residual magnitude limit $\zeta^{*}$. Each scenario refers to different initial and target object poses.}
\label{tab:various_object_experiment}
\end{table*}

We have designed three domains to evaluate the performance of our method as shown in Figure~\ref{fig:overall_illustration}. The first domain is the \textit{card domain}, where for each training episode, we uniformly sample the initial and the goal object positions on the table while its orientation along the z-axis is uniformly sampled in the range of $[0,2\pi)$. The second domain is the \emph{bump domain}. %
The object is initially placed on the right side of the bump, and the robot's task is to push the object over the bump and reorient it to match a goal position and orientation, which is uniformly sampled on top of the bump or the opposite side of the table. The initial and the goal orientations along the roll and pitch axis 
are uniformly sampled from the set [0, $\pi/2$, $\pi$, $3\pi/2$], while that along the yaw axis is uniformly sampled from $[0,2\pi)$. The third domain is the \emph{wall domain}. 
The initial pose of the object is fixed to the up-right pose near the wall.  The goal pose of the object is uniformly sampled on the left side of the top of the wall. We do not try to match the orientation in this domain. During training, we fix the density of the object. The default density of the object is 457.1\unit{\kilo\gram/\cubic\metre}, 200\unit{\kilo\gram/\cubic\metre}, and 150\unit{\kilo\gram/\cubic\metre} for the card, bump, and wall domain respectively. 

We use IssacGym~\cite{makoviychuk2021isaac} to train the policies, and use different policies for different domains. In the simulator, we run $24,576$ environments simultaneously. We use $C_1=0.02$, $C_2=1000$, $C_3=0.03$, $C_{4}=-100$, $\bar{d}=0.01\unit{\metre}$ and $\bar{\theta}=0.1\unit{\radian}$ for the reward function. 
We run $\pi_{post}$ to output actions at a frequency of 10Hz, while the controller runs at a frequency of 100Hz. For the action scale schedule, we use $N_{s}=10$, $\zeta_{o}=(0.06,0.1)$, and $\zeta^{*}=(0.02,0.03)$. 

For the real-world experiment, we use a RealSense D435 camera to obtain RGB images. We use RRT*~\cite{karaman2011sampling} to plan a trajectory from the initial robot configuration given by $\pi_{pre}^{S}$, and use Polymetis~\cite{Polymetis2021} to control the robot. We run $\pi_{post}$ outputs its action at 10Hz, and the controller runs at 500Hz. To address safety issues, we set the minimum damping ratio for $\rho[t]$ at 0.5. We wrapped the gripper of the robot with a glove that has a higher friction coefficient to match the setting of the simulation.  

\subsection{Results}
In this paper, we make the following claims. 
\begin{itemize}
    \item Claim 1: Jointly training $\pipre$ with $\pipost$ facilitates more efficient exploration than using a single policy with contact-inducing reward.
    \item Claim 2: Residual curriculum enables efficient exploration and learning of a policy that meets joint limits.
    \item Claim 3: Inverse differential kinematics and joint position controller is better suited for sim-to-real transfer on robots with large gear ratios, such as Panda.
    \item Claim 4: Our system can transfer to the real world and generalize to objects that were not seen during training.
\end{itemize}

To support claim 1, we compare our method against the benchmark that does not use $\pi_{pre}$ in the bump domain where we initialize the joint position of the robot in two different ways, as shown in Figure~\ref{fig:EE-ABOVE}. In both setups, the robot needs to learn that it
needs to contact the object on the right side to push it over the bump. Figure~\ref{fig:plot1} shows the results. We can see the effect of using $\pipre$: our method reaches
95\% success rate at around 6 billion steps, while the one without $\pipre$ and $\pipost$ decomposition fails to reach the same level of success rate. The policy in EE-AT-RIGHT setup does a little better than the one in EE-ABOVE, because the robot begins at a configuration near the right side of the object. However, it is much less data-efficient than our method.

To support claim 2, We compare methods with and without end-effector residual scaling during training. For the one without the end-effector residual, we fix the residual magnitude limit to $\zeta^{*}$ and train the policy. Figures~\ref{fig:plot2}, \ref{fig:plot3}, and \ref{fig:plot4} show the result in three domains. As the plot shows, except for the card domain, the one without scheduling fails to learn the task, demonstrating the importance of our scheduling scheme. For the card domain, We further compare our policy with the policy without scheduling in the real world. The policy without scheduling achieved 73\% success rate, and ours achieved 87\% as shown in Table~\ref{tab:various_object_experiment}. This shows that action scaling not only helps with exploration, but also in closing the sim-to-real gap.

To support claim 3, we compare our method against the method in~\cite{vices2019} where the policy outputs the next end-effector pose and the gains for an OSC in the real world. We use the same action scheduling and policy architecture as our system but use a different action space and controller. In the simulation, the OSC-based method achieves over 90\% success rates in the card and bump domains and over 80\% in the wall domain in simulation. We then compare them in the real world with the object that has the same physical parameters as the one used in training. Table~\ref{tab:various_object_experiment} shows the result. It shows that the OSC-based approach does much worse in the real world and achieves 13\%, 7\%, and 0\% success rates in card, bump, and wall domains respectively, while ours achieve 87\% success rates in all three. This indicates that OSC-based action space has a larger sim-to-real gap than our action space.

To support claim 4, we test our policy on 7 different objects in each domain. The objects vary in their mass, surface friction, scale, and deformation. We paint the objects so that it has the same visual appearance as the training data. The list of the objects and the results are shown in Table~\ref{tab:various_object_experiment}. For the default object that has the same physical properties as the one we used in training, we achieve 87\% success rate, indicating that our system can transfer to the real world. Furthermore, our method succeeds even for some objects whose physical properties (ex. non-rigid, heavy, etc.) are outside of our domain randomization range. The object with the lowest success rate is the water tissue box, which is highly non-rigid.

\section{Limitation}
Our current work is limited to fixed environments and objects, and must train new policies if they change. The future work would involve generalizing across different object shapes by incorporating shape information.

\small{
\section{Acknowledgement}
This work was supported by Institute of Information \& communications Technology Planning \& Evaluation (IITP) grant funded by the Korea government(MSIT) (No.2019-0-00075, Artificial Intelligence Graduate School Program(KAIST)), (No.2022-0-00311, Development of Goal-Oriented Reinforcement Learning Techniques for Contact-Rich Robotic Manipulation of Everyday Objects), (No. 2022-0-00612, Geometric and Physical Commonsense Reasoning based Behavior Intelligence for Embodied AI), and Samsung Electronics.
}
\addtolength{\textheight}{-1cm}   %

\bibliographystyle{IEEEtran}
\bibliography{root.bbl}

\begin{thebibliography}{10}
\providecommand{\url}[1]{#1}
\csname url@rmstyle\endcsname
\providecommand{\newblock}{\relax}
\providecommand{\bibinfo}[2]{#2}
\providecommand\BIBentrySTDinterwordspacing{\spaceskip=0pt\relax}
\providecommand\BIBentryALTinterwordstretchfactor{4}
\providecommand\BIBentryALTinterwordspacing{\spaceskip=\fontdimen2\font plus
\BIBentryALTinterwordstretchfactor\fontdimen3\font minus
  \fontdimen4\font\relax}
\providecommand\BIBforeignlanguage[2]{{%
\expandafter\ifx\csname l@#1\endcsname\relax
\typeout{** WARNING: IEEEtran.bst: No hyphenation pattern has been}%
\typeout{** loaded for the language `#1'. Using the pattern for}%
\typeout{** the default language instead.}%
\else
\language=\csname l@#1\endcsname
\fi
#2}}

\bibitem{gilwoo2015iros}
G.~Lee, T.~Lozano-Perez, and L.~P. Kaelbling, ``{Hierarchical Planning for
  Multi-Contact Non-Prehensile Manipulation},'' in \emph{International
  Conference on Intelligent Robots and Systems}, 2015.

\bibitem{jenny2013icra}
J.~Barry, T.~Lozano-Per\'{e}z, and L.~P. Kaelbling, ``{A Hierarchical Approach
  to Manipulation with Diverse Actions},'' in \emph{International Conference on
  Robotics and Automation}, 2013.

\bibitem{cheng2022contact}
X.~Cheng, E.~Huang, Y.~Hou, and M.~T. Mason, ``{Contact Mode Guided Motion
  Planning for Quasidynamic Dexterous Manipulation in 3D},'' in
  \emph{International Conference on Robotics and Automation}, 2022.

\bibitem{posa2014direct}
M.~Posa, C.~Cantu, and R.~Tedrake, ``{A Direct Method for Trajectory
  Optimization of Rigid Bodies Through Contact},'' \emph{The International
  Journal of Robotics Research}, 2014.

\bibitem{mordatch2012contact}
I.~Mordatch, Z.~Popovi{\'c}, and E.~Todorov, ``{Contact-Invariant Optimization
  for Hand Manipulation},'' in \emph{ACM SIGGRAPH/Eurographics Symposium on
  Computer Animation}, 2012.

\bibitem{bernardo2020rss}
B.~Aceituno-Cabezas and A.~Rodriguez, ``{A Global Quasi-Dynamic Model for
  Contact-Trajectory Optimization},'' in \emph{Robotics: Science and Systems},
  2020.

\bibitem{lynchcontact}
I.~Kao, K.~M. Lynch, and J.~W. Burdick, ``{Contact Modeling and
  Manipulation},'' in \emph{Springer Handbook of Robotics}, 2016.

\bibitem{miyazawaISATP2005}
K.~Miyazawa, Y.~Maeda, and T.~Arai, ``{Planning of Graspless Manipulation Based
  on Rapidly-Exploring Random Trees},'' in \emph{Assembly and Task Planning:
  From Nano to Macro Assembly and Manufacturing}, 2005.

\bibitem{chen2022system}
T.~Chen, J.~Xu, and P.~Agrawal, ``{A System for General In-Hand Object
  Re-Orientation},'' in \emph{Conference on Robot Learning}, 2022.

\bibitem{chen2022visual}
T.~Chen, M.~Tippur, S.~Wu, V.~Kumar, E.~Adelson, and P.~Agrawal, ``{Visual
  Dexterity: In-Hand Dexterous Manipulation from Depth},'' \emph{arXiv}, 2022.

\bibitem{andrychowicz2020learning}
O.~M. Andrychowicz, B.~Baker, M.~Chociej, R.~Jozefowicz, B.~McGrew,
  J.~Pachocki, A.~Petron, M.~Plappert, G.~Powell, A.~Ray, \emph{et~al.},
  ``{Learning Dexterous In-Hand Manipulation},'' \emph{The International
  Journal of Robotics Research}, 2020.

\bibitem{hwangbo2019science}
J.~Hwangbo, J.~Lee, A.~Dosovitskiy, D.~Bellicoso, V.~Tsounis, V.~Koltun, and
  M.~Hutter, ``{Learning Agile and Dynamic Motor Skills for Legged Robots},''
  \emph{Science Robotics}, 2019.

\bibitem{lee2020learning}
J.~Lee, J.~Hwangbo, L.~Wellhausen, V.~Koltun, and M.~Hutter, ``{Learning
  Quadrupedal Locomotion over Challenging Terrain},'' \emph{Science Robotics},
  2020.

\bibitem{TAMPsurvey}
C.~R. Garrett, R.~Chitnis, R.~Holladay, B.~Kim, T.~Silver, L.~P. Kaelbling, and
  T.~Lozano-Perez, ``{Integrated Task and Motion Planning},'' \emph{Annual
  Review of Control, Robotics, and Autonomous Systems}, 2021.

\bibitem{chen2020learning}
D.~Chen, B.~Zhou, V.~Koltun, and P.~Kr{\"a}henb{\"u}hl, ``{Learning by
  Cheating},'' in \emph{Conference on Robot Learning}, 2020.

\bibitem{vices2019}
R.~Mart{\'\i}n-Mart{\'\i}n, M.~A. Lee, R.~Gardner, S.~Savarese, J.~Bohg, and
  A.~Garg, ``{Variable Impedance Control in End-Effector Space: An Action Space
  for Reinforcement Learning in Contact-Rich Tasks},'' in \emph{International
  Conference on Intelligent Robots and Systems}, 2019.

\bibitem{nakanishi2008operational}
J.~Nakanishi, R.~Cory, M.~Mistry, J.~Peters, and S.~Schaal, ``{Operational
  Space Control: A Theoretical and Empirical Comparison},'' \emph{The
  International Journal of Robotics Research}, 2008.

\bibitem{moura2022non}
J.~Moura, T.~Stouraitis, and S.~Vijayakumar, ``{Non-Prehensile Planar
  Manipulation via Trajectory Optimization with Complementarity Constraints},''
  in \emph{International Conference on Robotics and Automation}, 2022.

\bibitem{zito2012two}
C.~Zito, R.~Stolkin, M.~Kopicki, and J.~L. Wyatt, ``{Two-Level RRT Planning for
  Robotic Push Manipulation},'' in \emph{International Conference on
  Intelligent Robots and Systems}, 2012.

\bibitem{liang2022learning}
J.~Liang, X.~Cheng, and O.~Kroemer, ``{Learning Preconditions of Hybrid
  Force-Velocity Controllers for Contact-Rich Manipulation},'' in
  \emph{Conference on Robot Learning}, 2022.

\bibitem{akkaya2019solving}
I.~Akkaya, M.~Andrychowicz, M.~Chociej, M.~Litwin, B.~McGrew, A.~Petron,
  A.~Paino, M.~Plappert, G.~Powell, R.~Ribas, \emph{et~al.}, ``{Solving Rubik's
  Cube with a Robot Hand},'' \emph{arXiv}, 2019.

\bibitem{allshire2021transferring}
A.~Allshire, M.~Mittal, V.~Lodaya, V.~Makoviychuk, D.~Makoviichuk, F.~Widmaier,
  M.~W{\"u}thrich, S.~Bauer, A.~Handa, and A.~Garg, ``{Transferring Dexterous
  Manipulation from GPU Simulation to a Remote Real-World Trifinger},''
  \emph{arXiv}, 2021.

\bibitem{dextreme2022}
A.~Handa, A.~Allshire, V.~Makoviychuk, A.~Petrenko, R.~Singh, J.~Liu,
  D.~Makoviichuk, K.~Van~Wyk, A.~Zhurkevich, B.~Sundaralingam, \emph{et~al.},
  ``{DeXtreme: Transfer of Agile In-Hand Manipulation from Simulation to
  Reality},'' \emph{arXiv}, 2022.

\bibitem{yuan2018rearrangement}
W.~Yuan, J.~A. Stork, D.~Kragic, M.~Y. Wang, and K.~Hang, ``{Rearrangement with
  Nonprehensile Manipulation Using Deep Reinforcement Learning},'' in
  \emph{International Conference on Robotics and Automation}, 2018.

\bibitem{yuan2019end}
W.~Yuan, K.~Hang, D.~Kragic, M.~Y. Wang, and J.~A. Stork, ``{End-to-End
  Nonprehensile Rearrangement with Deep Reinforcement Learning and
  Simulation-to-Reality Transfer},'' \emph{Robotics and Autonomous Systems},
  2019.

\bibitem{lowrey2018reinforcement}
K.~Lowrey, S.~Kolev, J.~Dao, A.~Rajeswaran, and E.~Todorov, ``{Reinforcement
  Learning for Non-Prehensile Manipulation: Transfer from Simulation to
  Physical System},'' in \emph{International Conference on Simulation,
  Modeling, and Programming for Autonomous Robots}, 2018.

\bibitem{peng2018sim}
X.~B. Peng, M.~Andrychowicz, W.~Zaremba, and P.~Abbeel, ``{Sim-to-Real Transfer
  of Robotic Control with Dynamics Randomization},'' in \emph{International
  Conference on Robotics and Automation}, 2018.

\bibitem{zhou2022learning}
W.~Zhou and D.~Held, ``{Learning to Grasp the Ungraspable with Emergent
  Extrinsic Dexterity},'' in \emph{Conference on Robot Learning}, 2022.

\bibitem{schulman2017proximal}
J.~Schulman, F.~Wolski, P.~Dhariwal, A.~Radford, and O.~Klimov, ``{Proximal
  Policy Optimization Algorithms},'' \emph{arXiv}, 2017.

\bibitem{zhou2019continuity}
Y.~Zhou, C.~Barnes, J.~Lu, J.~Yang, and H.~Li, ``{On the Continuity of Rotation
  Representations in Neural Networks},'' in \emph{Conference on Computer Vision
  and Pattern Recognition}, 2019.

\bibitem{s3k2021}
M.~Vecerik, J.-B. Regli, O.~Sushkov, D.~Barker, R.~Pevceviciute,
  T.~Roth{\"o}rl, R.~Hadsell, L.~Agapito, and J.~Scholz, ``{S3K:
  Self-Supervised Semantic Keypoints for Robotic Manipulation via Multi-View
  Consistency},'' in \emph{Conference on Robot Learning}, 2021.

\bibitem{moco2020}
K.~He, H.~Fan, Y.~Wu, S.~Xie, and R.~Girshick, ``{Momentum Contrast for
  Unsupervised Visual Representation Learning},'' in \emph{Conference on
  Computer Vision and Pattern Recognition}, 2020.

\bibitem{oord2018representation}
A.~v.~d. Oord, Y.~Li, and O.~Vinyals, ``{Representation Learning with
  Contrastive Predictive Coding},'' \emph{arXiv}, 2018.

\bibitem{jiang2022vima}
Y.~Jiang, A.~Gupta, Z.~Zhang, G.~Wang, Y.~Dou, Y.~Chen, L.~Fei-Fei,
  A.~Anandkumar, Y.~Zhu, and L.~Fan, ``{VIMA: General Robot Manipulation with
  Multimodal Prompts},'' \emph{arXiv}, 2022.

\bibitem{cmaes2003}
N.~Hansen, S.~D. M{\"u}ller, and P.~Koumoutsakos, ``{Reducing the Time
  Complexity of the Derandomized Evolution Strategy with Covariance Matrix
  Adaptation (CMA-ES)},'' \emph{Evolutionary Computation}, 2003.

\bibitem{makoviychuk2021isaac}
V.~Makoviychuk, L.~Wawrzyniak, Y.~Guo, M.~Lu, K.~Storey, M.~Macklin,
  D.~Hoeller, N.~Rudin, A.~Allshire, A.~Handa, \emph{et~al.}, ``{Isaac Gym:
  High Performance GPU-Based Physics Simulation for Robot Learning},''
  \emph{arXiv}, 2021.

\bibitem{karaman2011sampling}
S.~Karaman and E.~Frazzoli, ``{Sampling-Based Algorithms for Optimal Motion
  Planning},'' \emph{The International Journal of Robotics Research}, 2011.

\bibitem{Polymetis2021}
{Lin, Yixin and Wang, Austin S. and Sutanto, Giovanni and Rai, Akshara and
  Meier, Franziska}, ``Polymetis,'' 2021.

\end{thebibliography}
\end{document}